# Understanding Neural Pathways in Zebrafish through Deep Learning and High Resolution Electron Microscope Data

Ishtar Nyawĩra
Pittsburgh Supercomputing Center
Carnegie Mellon University
University of Pittsburgh
300 South Craig St.
Pittsburgh, PA  15213
1 (412) 268-4960
ishtarnyawira@pitt.edu

Kristi Bushman
Pittsburgh Supercomputing Center
Carnegie Mellon University
University of Pittsburgh
300 South Craig St.
Pittsburgh, PA  15213
1 (412) 268-4960
k.bushman@pitt.edu

Iris Qian
Pittsburgh Supercomputing Center
Carnegie Mellon University
300 South Craig St.
Pittsburgh, PA  15213
1 (412) 268-4960
liyunshq@andrew.cmu.edu

Annie Zhang
Pittsburgh Supercomputing Center
Carnegie Mellon University
300 South Craig St.
Pittsburgh, PA  15213
1 (412) 268-4960
anniez@andrew.cmu.edu

## ABSTRACT

The tracing of neural pathways through large volumes of image data is an incredibly tedious and time-consuming process that significantly encumbers progress in neuroscience. We are exploring deep learning's potential to automate segmentation of high-resolution scanning electron microscope (SEM) image data to remove that barrier. We have started with neural pathway tracing through 5.1GB of whole-brain serial-section slices from larval zebrafish collected by the Center for Brain Science at Harvard University. This kind of manual image segmentation requires years of careful work to properly trace the neural pathways in an organism as small as a zebrafish larva (approximately 5mm in total body length). In automating this process, we would vastly improve productivity, leading to faster data analysis and breakthroughs in understanding the complexity of the brain. We will build upon prior attempts to employ deep learning for automatic image segmentation extending methods for unconventional deep learning data.

## 1   INTRODUCTION

In recent years, Convolutional Neural Networks (CNNs) have had extensive success in pixel-wise image segmentation. While these networks have been applied to facial image recognition [8] and street-view object recognition [1], we have yet to find broad use of convolutional neural networks for the kind of nontraditional biomedical image data segmentation that is often used in connectomics. Connectomics, the building of complete mappings of the neural pathways in an organism (connectomes), has great need for more efficient data creation and analysis. It's worth noting that the primary differences in CNN topologies for biomedical image data lies in the image resolution and contrast. Network architectures successful for tasks such as street-view object recognition generally operate on comparatively low-resolution, high-contrast images, whereas SEM images are of much higher resolution and lower contrast. Additionally, data annotations collected to build connectomes often consist of single-pixel points in the center of neurons (Figs. 1a and 1b). Currently, the collection of training data necessary to build connectomes requires years of meticulous work as well as the need for constant updating as more information is gathered.

Our datasets focus on zebrafish larvae. Although these are very small organisms (5mm in length on average), large amounts of data must be collected to properly map their biological structure. The amount of data grows proportionally to brain volume, and for larger organisms it is beyond today's storage systems. To put this in perspective, a mouse brain is on average 430mm$^3$ in size, which would (if fully imaged) amount to approximately 1.4 exabytes of raw imaging data [3]. This kind of data is far too large for humans to efficiently annotate manually. Thus, we are working to automate this process. We chose convolutional neural networks with specialized data augmentation methods (namely, flood-fill and region growing operations) to increase efficiency and free neuroscientists and neuroscience students from this tedious and relatively mechanical task. Our preliminary findings are encouraging, indicating a potentially simplified a process vital to the exploration of neuroscience so that neuroscientists may better apply their expertise to data analysis and biomedical expedition.

## 2 DATA & METHODS

We received 5.1GB of SEM image data (4900 images, each 1024×1024 pixels in size) from the Center for Brain Science at Harvard. However, we were able to multiply that data to 61.2GB after applying several transformations for image augmentation – including flipping, rotating, and translating images – as is often necessary for the training of a neural network. Our datasets were compiled and consumed by our neural networks with the use of PSC's state-of-the-art Bridges supercomputer (https://dl.acm.org/citation.cfm?id=2792775), distinguished for its flexibility and scalable, converged approach to high-performance computing, artificial intelligence, and big data. We accelerated our computations using Bridges' NVIDIA Tesla K80 and P100 GPUs.

The labeled data we received poses a challenge for deep learning. At the center of every identified neuron, a single-pixel click-point was marked (Fig. 1b). However, deep learning algorithms typically rely heavily on shape recognition to effectively learn how to annotate images fed into the network. Thus, due to the point nature of our labeled data, we had to grow the annotated regions to fill out the entire neuron. We did this in two automated ways – flood fill and seeded region growing – which resulted in varying degrees of success after our networks were trained. Both are detailed in code linked from their respective references.

Our first method was to use a flood-fill operation [9], which fills shapes from their click-points to their borders. This requires the neurons to have clearly defined borders that are completely closed (a single pixel gap can cause leaking). Thus, the first step was to threshold our greyscale images to binary so that we can focus on only two pixel values and thus only two classes (background and borders). Next, a series of morphological operations were applied to help close any gaps in the borders. The first of which was used to isolate the image skeleton, which simplifies much of the image complexity. The second operation in the flood-fill algorithm morphologically closed gaps in the skeleton, creating closed outlined shapes with defined borders for our next operation. That final operation flood-filled the outlined neurons to their borders, starting from their respective click-point centers. This method was very sensitive to parameter choice, especially to changes made to the radius of the structuring element used in the closing operation. Too large a radius, and the flood-fill method would be inflexible and more prone to missing small neurons. Too small a radius, and the method would be prone to spilling out over borders.

Our second method of growing the annotated regions was called seeded region growing [7]. This method was more involved, but ultimately produced better results. The method works to iteratively grow a region from a single seed pixel by comparing all neighboring pixels to the current region (starting with the initial single-pixel region). The measure of similarity used is simply the difference between a pixel's intensity value and the region's mean intensity value. With every iteration, the neighboring pixel with the smallest difference is added to the region and the region's mean intensity value is recalculated. This process stops when the similarity measure between the region mean and all neighboring pixels becomes larger than a given threshold. The region growing method was much less parameter sensitive than flood-fill.

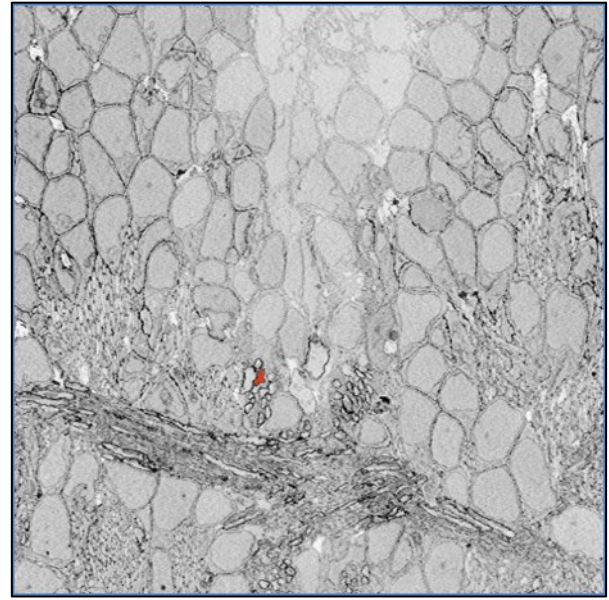

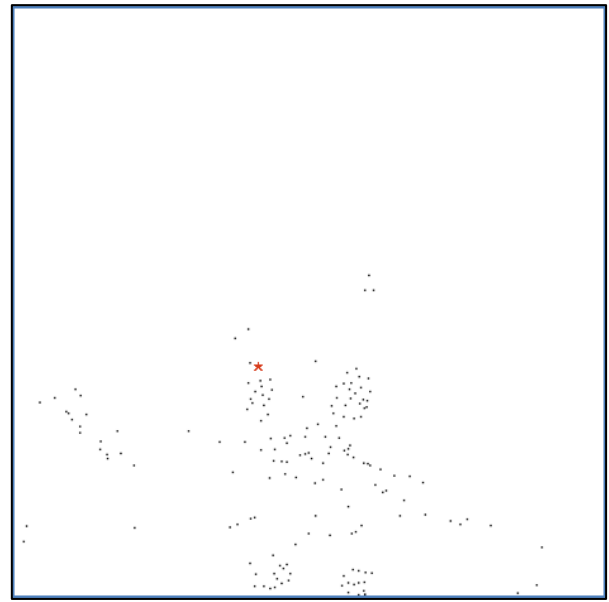

**Figures 1a & 1b: SEM Core-Brain Serial Section of Zebrafish from the head of the zebrafish. Original SEM image (top) and its annotated counterpart (bottom), slice 4500 of 4900 [5]. Red spots tie a neuron to its click-point location.**

Both methods fill neurons, but to different degrees of success. We noted that the flood-fill operation would sometimes miss neurons whose membranes were thin or weren't darkly dyed, and often neurons were not being filled completely. Many filled-in shapes had a boxy appearance that did not match the contour of the neuron. Our region growing alternative did a better job of not missing neurons and filling them completely. This method also followed the contours of the neurons to a much better degree. It follows that the more accurately the neuron is filled, the easier it is for a deep neural net to learn to recognize and annotate the shapes of neurons.

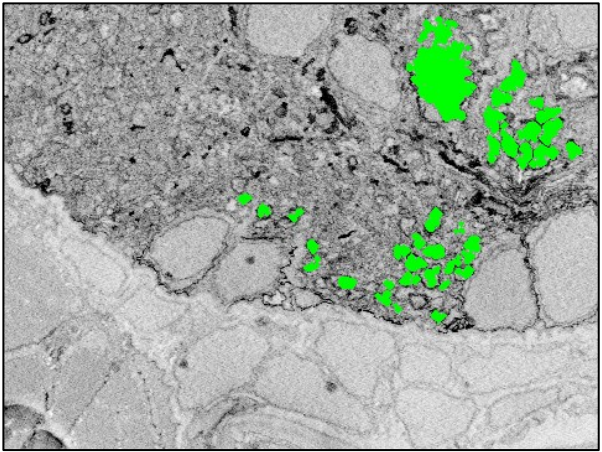

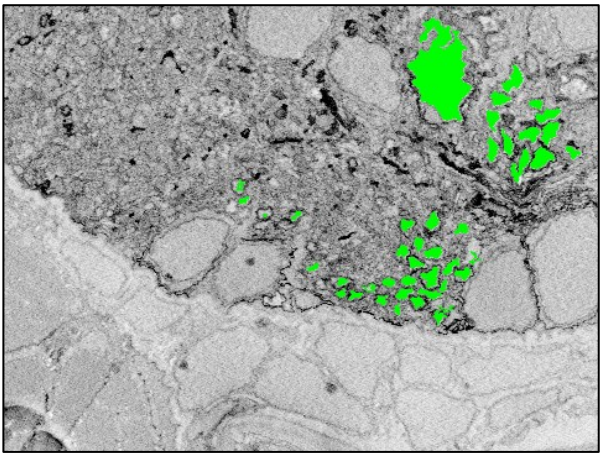

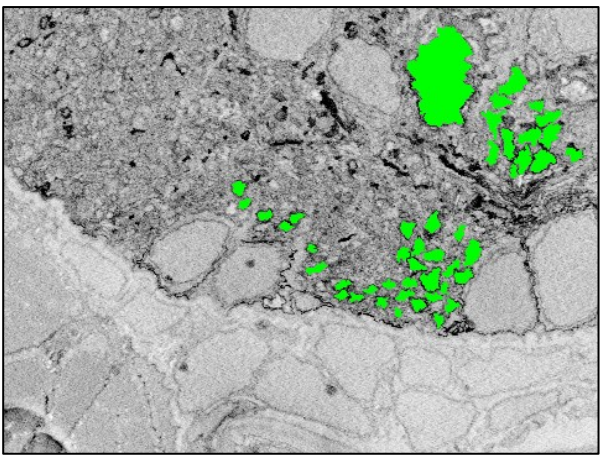

**Figures 2a & 2b & 2c: Flood-fill (top), region-growing (middle), and manual (bottom) annotations.**

Although we had some success with the seeded region growing method, our results were still limited by inaccuracies in this step. The region-growing method, although better than the flood-fill method, could still on occasion miss neurons with especially thin or light borders or have difficulty filling them completely. We determined that it was necessary to segment some of our images manually to be able to validate our previous results. We also suspect that we could get even better results by training our nets on these higher-quality segmentation masks. Unfortunately, manual segmentation is a very time-consuming task, which is why it served as the motivation for this research. For high quality annotations, segmentation takes forty-five minutes to one hour per image to complete. We currently have enough images segmented to be able to validate our previous results and train a net with a small amount of data.

## 2.1 Software & Frameworks

We determined that TensorFlow, an open-source framework developed by Google, provided a sufficient variety of open source neural network architectures and the kind of plasticity we needed to mutate those architectures and their hyperparameters to address our particular datasets.

We have worked closely with SetNet [1] and UNet [2], two neural networks already applied successfully to pixel-wise image classification, in hopes of manipulating them for our unique needs. And while Tseng Kuan-Lun's TensorFlow version of SegNet [5] was efficient and complete, we used Marko Jocić's Keras implementation of UNet [6] because the TensorFlow version was not fully functional.

We modified Kuan-Lun's code to include an unpooling layer, as described in the SegNet paper. The TensorFlow APIs do not currently include an unpooling method, so deconvolution was used in Kuan-Lun's code instead. We replaced these deconvolution layers with unpooling layers, using code found in the issues section of TensorFlow on GitHub [13].

We are continuing work to discover the best way to effectively modify the SegNet and UNet architectures and hyperparameters in accordance with finding the best way to annotate our datasets. With this kind of careful attention to detail, we can ensure that we are taking the appropriate steps in forwarding our progress and discovering the advantages of deep learning in path tracing through large volumes of image data.

## 3 EVALUATION METRICS

To evaluate our results, we used standard measures of accuracy that employ comparisons between true positives (TP), false positives (FP), true negatives (TN), and false negatives (FN), where positive refers to our neuron class. We chose to forgo the use of a general accuracy measure (Equation 1) because it is very sensitive to the imbalance of classes, there being far more background pixels than neuron pixels:

$$\text{ACC} = \frac{\text{TP+TN}}{\text{All}} \qquad (1)$$

Additionally, we chose to avoid the Intersection over Union (IoU) (Equation 2), also known as the Jaccard Index. This is because, although it does not consider TN (accurately classified background pixels, i.e. the majority of pixels), it is still sensitive to class imbalances and is not robust against chance agreement between samples:

$$\text{JAC} = \frac{\text{TP}}{\text{TP+FP+FN}} \qquad (2)$$

Note that originally, UNet makes use of the Dice Coefficient (Equation 3), a measure only slightly different from the Jaccard Index in that both eliminate TN from the numerator and denominator and both do not supply chance adjustment:

$$DICE = \frac{2TP}{2TP+FP+FN} = \frac{2JAC}{1+JAC} \quad (3)$$

Instead we chose to rely on measures that more accurately reflect the quality of neuron segmentation, namely: the Cohen Kappa Coefficient (Equation 4) and the Area Under the Receiver Operating Characteristic Curve or AUROC (Equation 7):

$$KAP = \frac{Pa-Pc}{1-Pc} = \frac{fa-fc}{N-fc} \quad (4)$$

$$fa = TP + TN \quad (5)$$

$$fc = \frac{(TN+FN)*(TN+FP) + (FP+TP)*(FN+TP)}{N} \quad (6)$$

Where Pa is the probability of agreement between classes, Pc is the probability of chance agreement, and N is the total number of observations. Please note that because our TP, TN, FP, FN measures are provided as percentages, N is always 1.

$$AUROC = 1 - \frac{FPR+FNR}{2} = 1 - \frac{1}{2}\left(\frac{FP}{FP+TN} + \frac{FN}{FN+TP}\right) \quad (7)$$

The Cohen Kappa Coefficient simply measures the agreement between two samples. We decided to use it because when the segmentations have a high class imbalance, it's best to use metrics with chance adjustment (which adjusts for the times when agreement between two samples could be by chance). While TN is considered in this metric, chance adjustment decreases its impact on the results [10]. The adjusted rand index measure takes these same precautions, but we chose to use the Kappa measure because of its simplicity.

There are some guidelines provided for the evaluation of a resulting Kappa Coefficient. We will use the guidelines provided by Landis and Koch [12]. It's important to note that these are relatively arbitrary guidelines and thus, naturally, universal agreement on their use does not exist.

**Table 1: Landis & Koch Kappa Coefficient Guidelines**

| Value | Level of Agreement |
|---|---|
| < 0 | no agreement |
| 0–0.20 | slight agreement |
| 0.21–0.40 | fair agreement |
| 0.41–0.60 | moderate agreement |
| 0.61–0.80 | substantial agreement |
| 0.81–1 | almost perfect agreement |

The second quantitative measure of accuracy we chose to use is the Area Under the ROC Curve (AUROC), which measures the probability that a randomly chosen positive pixel is deemed to have a higher probability of being positive than a randomly chosen negative pixel. It is important to note that the two measures used to plot AUROC, sensitivity (TP rate or recall) and fall-out (FP rate), are not often used to evaluate medical image data. This is because they are sensitive to class imbalance and are thus prone to penalizing errors in small segments more so than errors in large segments [10]. This results in very low values for both sensitivity and fall-out.

However, because the ROC plots these two measures against one another, the focus is shifted to the way the two measures relate to one another rather than on what the values themselves are. We found this was a helpful way to quantitatively assess our data's ability to classify the positive neuron class.

There are some general guidelines for evaluating AUROC, which follow a traditional academic point system [11]:

**Table 2: Traditional Academic AUROC Curve Guidelines**

| Value | Level of Agreement |
|---|---|
| 0.50–0.60 | no agreement (F) |
| 0.60–0.70 | poor agreement (D) |
| 0.70–0.80 | fair agreement (C) |
| 0.80–0.90 | good agreement (B) |
| 0.90–1 | excellent agreement (A) |

## 4 RESULTS

The quantitative results we were able to produce are captured in Tables 3 and 4. The prediction images each net produced and their respective original and true masks (our qualitative results) are found in Figs. 3, 4, and 5.

The values in Tables 3 and 4 were calculated using models trained on 35,000 and 3,838 training images (for flood-fill/region-growing and manual annotation respectively) and a test set of 50 randomly-selected images. Our test set consisted of manually annotated segmentations, since that is the closest that we have to a gold-standard annotation. The Kappa measure and AUROC curve were calculated using Equations 4 and 7, respectively.

**Table 3: UNet's Kappa Coefficient & AUROC Results w. Varying Net & Data Annotations**

|  | Kappa Coefficient | AUROC |
|---|---|---|
| **Flood-Fill** (35,000 images) | 0.674656 | 0.835366 |
| **Region Growing** (35,000 images) | 0.664252 | 0.775311 |
| **Manual Annotation** (3,838 images) | 0.508445 | 0.818312 |

Table 4: SegNet's Kappa Coefficient & AUROC Results w. Varying Net & Data Annotations

|  | Kappa Coefficient | AUROC |
| --- | --- | --- |
| **Flood-Fill** (35,000 images) | 0.428534 | 0.816803 |
| **Region Growing** (35,000 images) | 0.557270 | 0.896916 |
| **Manual Annotation** (3,838 images) | 0.615110 | 0.965385 |

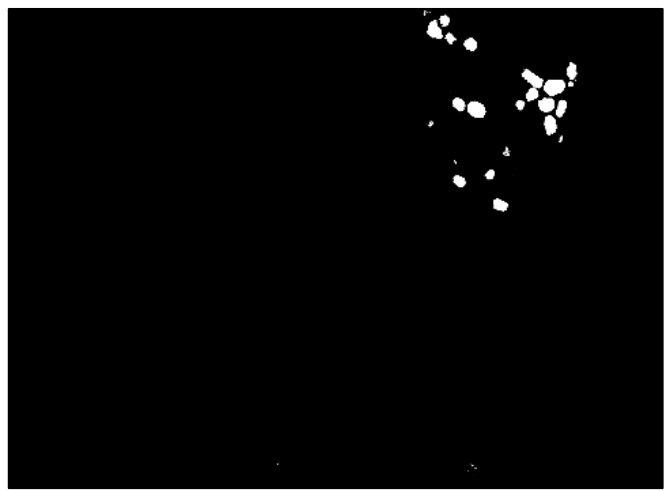

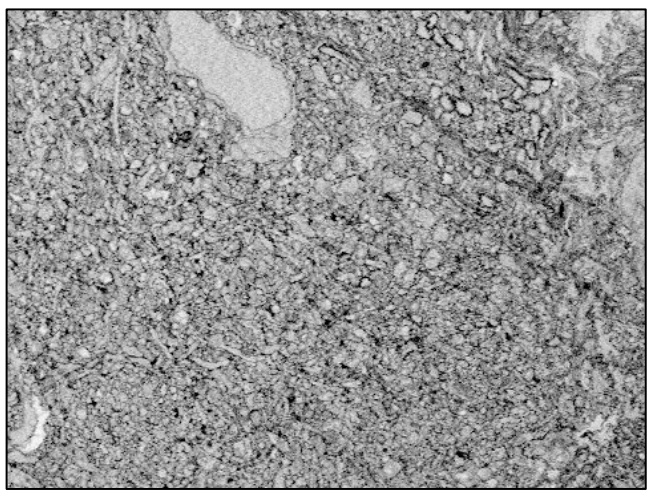

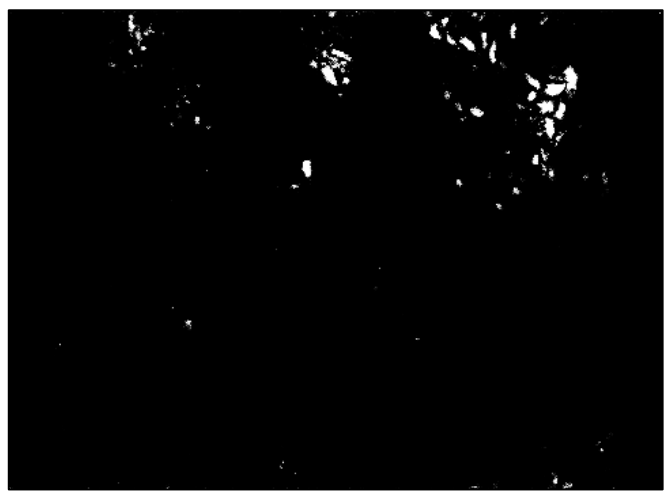

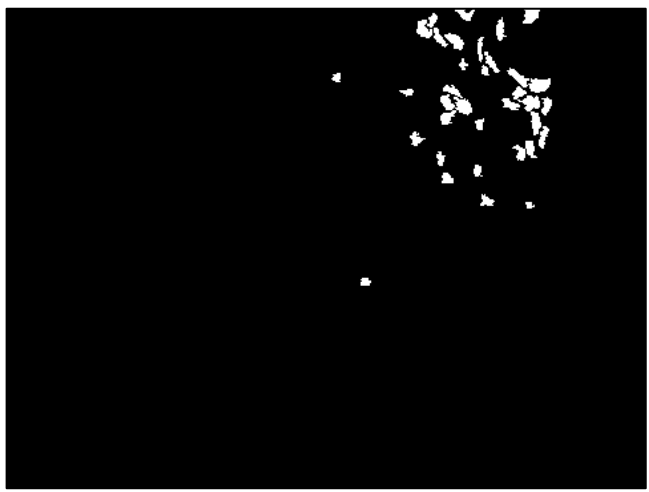

**Figures 3a & 3b:** The original zebrafish slice (top) and manually annotated label (bottom) from image 1930 of 4500. White spots are neuron segmentations.

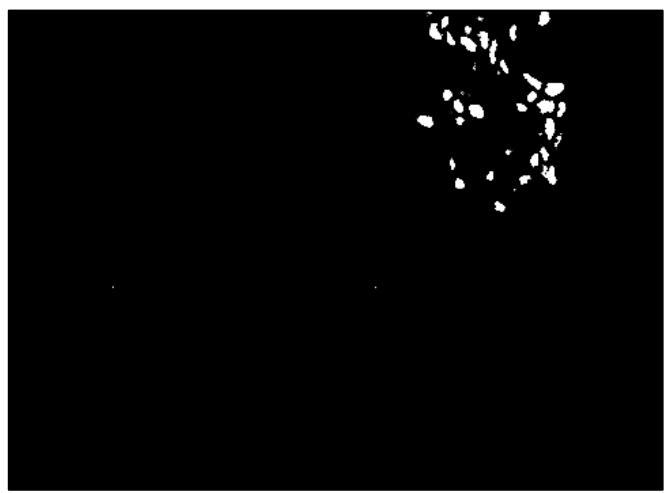

**Figures 4a & 4b & 4c:** UNet predicted results from image 1930 of 4500 using flood-fill (top), manual (middle) and region growing (bottom). White spots are neuron segmentations.

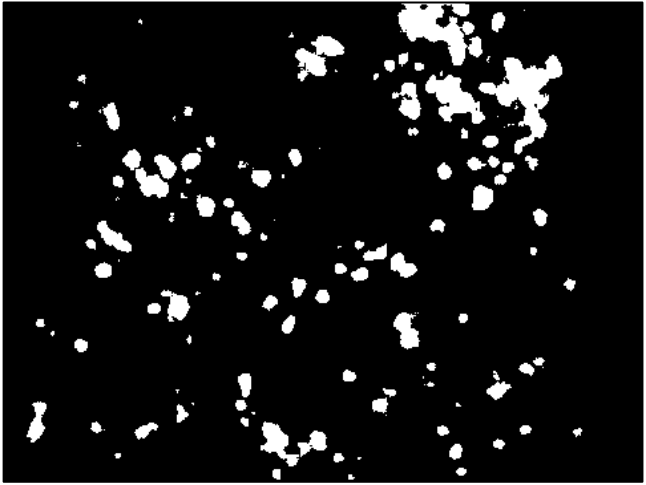

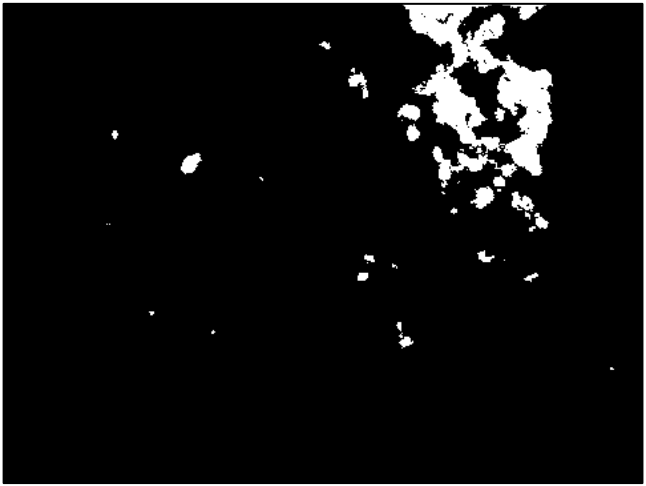

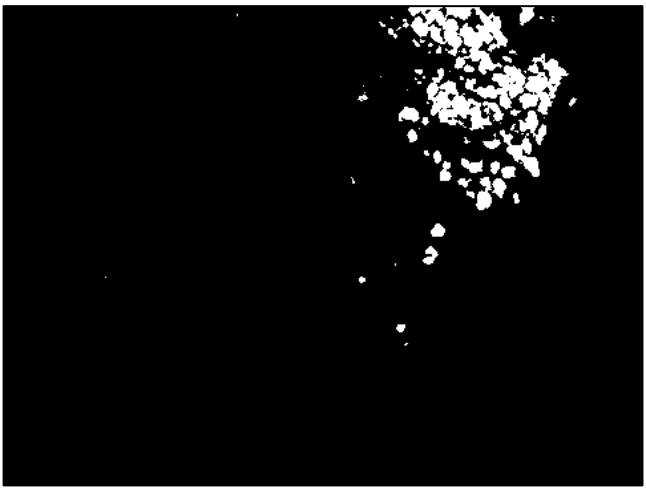

**Figures 5a & 5b & 5c: SegNet predicted results from image 1930 of 4500 using flood-fill (top), manual (middle) and region growing (bottom). White spots are neuron segmentations.**

## 4.1 Discussion of Results

Per the Landis and Koch Kappa Coefficient Guidelines (Table 1), UNet's flood-fill and region-growing results displayed substantial agreement between prediction and true mask samples, while its manual annotation results showed moderate agreement. SegNet's flood-fill and region-growing Kappa measure results displayed moderate agreement, while the manual annotation results depicted substantial agreement.

We suspected this could be because UNet relies on a large training set size to produce a good prediction model. On the other hand, it appears that with our data SegNet is just the opposite, depending more on segmentation quality than on training set size. This could be substantiated by the fact that the original SegNet paper had a dataset of only 367 training images, and was thus built to use a smaller dataset. We could not find the exact size of UNet's original training set size; however, we will note that the authors used "extensive data augmentation" in order to grow the small ISBI datasets they used [2].

Additionally, because the Kappa Coefficient is well suited for data with class imbalances, it's able to detect that UNet is generally more sensitive to small segments than SegNet and is better able to segment them individually. SegNet, on the other hand, is able to detect the general area of neurons, but struggles to define individual neurons.

Evaluation of the AUROC Curve is much clearer, although perhaps not initially. It is important to note that while UNet has higher Kappa Coefficient results in general, SegNet has higher AUROC scores on average. Additionally, SegNet's AUROC and Kappa measure scores appear to be much more tightly correlated than UNet's.

We believe this is because UNet had a higher tendency to label positive pixels as negative (a higher FN rate) than SegNet, especially for the region growing and manually annotated data (see Appendix A). And while UNet's FP rate is lower, it is not enough to offset its much higher FN rate (Equation 7).

## 5 CONCLUSION

In recent years, deep learning has been effectively utilized in the pixel-wise segmentation of image data. This was largely motivated by the need to automate time consuming and human-intensive tasks. Much of this has relied upon deep neural networks' ability to learn features though edge and shape recognition. In the field of connectomics, however, we must use novel methods in order to make effective use of data that is unconventional for deep learning.

Although we trained SegNet with approximately 90% less data for the manual annotations compared to the flood-fill and region-growing annotations, the resulting Kappa Coefficient and AUROC results for its manually annotated training set is slightly higher. There is room for improvement regarding SegNet's ability to more precisely annotate the borders and contours of neurons. However, we believe that with more manually annotated data, these results will improve further, both quantitatively and qualitatively.

With UNet, we found the greatest room for improvement in its high FN rate. We suspect that in order to tackle this challenge, we would need to look into other class imbalance solutions. Still, UNet produced rather impressive prediction images that better fit the contours of the neuron shapes.

In this paper, we made use of atypical data, adapting it for deep network leaning and returned results that appeared promising. Presently, we are working to manually annotate more images and discover better methods of managing class imbalance in training data so that we may achieve even better results.

# A  SUPPLEMENTARY METRICS

This appendix functions to supplement the metrics used to analyze the results produced by both UNet and SegNet.

FNR = False Negative Rate (Negative = background)

FPR = False Positive Rate (Positive = neuron)

## A.1  SegNet Confusion Matrices & FNR & FPR

*A.1.1 Flood-Fill Confusion Matrix & FNR & FPR*

|                   | Predicted Background | Predicted Neuron |
|-------------------|----------------------|------------------|
| Actual Background | 0.9259               | 0.0421           |
| Actual Neuron     | 0.0104               | 0.0217           |

FNR = 0.322915, FPR = 0.043478

*A.1.2 Region Growing Confusion Matrix & FNR & FPR*

|                   | Predicted Background | Predicted Neuron |
|-------------------|----------------------|------------------|
| Actual Background | 0.9343               | 0.0336           |
| Actual Neuron     | 0.0055               | 0.0266           |

FNR = 0.171410, FPR = 0.034758

*A.1.3 Manual Annotation Confusion Matrix & FNR & FPR*

|                   | Predicted Background | Predicted Neuron |
|-------------------|----------------------|------------------|
| Actual Background | 0.9329               | 0.0351           |
| Actual Neuron     | 0.0011               | 0.0310           |

FNR = 0.032981, FPR = 0.036248

## A.2  UNet Confusion Matrices & FNR & FPR

*A.2.1 Flood-Fill Confusion Matrix & FNR & FPR*

|                   | Predicted Background | Predicted Neuron |
|-------------------|----------------------|------------------|
| Actual Background | 0.9579               | 0.0099           |
| Actual Neuron     | 0.0103               | 0.0219           |

FNR = 0.319044, FPR = 0.010224

*A.2.3 Region Growing Confusion Matrix & FNR & FPR*

|                   | Predicted Background | Predicted Neuron |
|-------------------|----------------------|------------------|
| Actual Background | 0.9648               | 0.0030           |
| Actual Neuron     | 0.0143               | 0.0178           |

FNR = 0.446302, FPR = 0.003077

*A.2.3 Manual Annotation Confusion Matrix & FNR & FPR*

|                   | Predicted Background | Predicted Neuron |
|-------------------|----------------------|------------------|
| Actual Background | 0.9402               | 0.0276           |
| Actual Neuron     | 0.0108               | 0.0214           |

FNR = 0.334855, FPR = 0.028521

# ACKNOWLEDGMENTS

This work used the Extreme Science and Engineering Discovery Environment (XSEDE), which is supported by National Science Foundation award number 1548562. Specifically, it used the Bridges system, which is supported by NSF award number 1445606, at the Pittsburgh Supercomputing Center (PSC). Support for student interns was provided by the Bridges project (NSF award 1445606) and the Commonwealth of Pennsylvania.

Special thanks to Nick Nystrom, Joel Welling, Paola Buitrago, John Urbanic, Arthur Wetzel, and Tamara Cherwin at PSC. Thanks also to Florian Engert, David Hildebrand, and their students at the Center for Brain Science at Harvard, who compiled and shared their zebrafish data.